\documentclass[letterpaper, 10 pt, conference]{ieeeconf}
\IEEEoverridecommandlockouts
\usepackage[usenames]{xcolor}
\usepackage{amsmath}
\usepackage{amssymb}
\usepackage{diagbox}
\usepackage{epsfig}
\usepackage{graphicx}
\usepackage{hyperref}
\usepackage{mathptmx}
\usepackage{multirow}
\usepackage{times}
\usepackage{url}

\newcolumntype{L}[1]{>{\raggedright\let\newline\\\arraybackslash\hspace{0pt}}m{#1}}
\newcolumntype{C}[1]{>{\centering\let\newline\\\arraybackslash\hspace{0pt}}m{#1}}
\newcolumntype{R}[1]{>{\raggedleft\let\newline\\\arraybackslash\hspace{0pt}}m{#1}}

\usepackage{ragged2e}

\usepackage{xifthen}
\usepackage{xparse}
\usepackage{subdepth}
\usepackage{leftidx}
\usepackage{bm}

\newcommand{\bbm}{\begin{bmatrix}}
\newcommand{\ebm}{\end{bmatrix}}

\DeclareMathAlphabet{\mbf}{OT1}{ptm}{b}{n}
\newcommand{\mbs}[1]{{\bm{#1}}}

\newcommand{\mbsbar}[1]{{\overline{\boldsymbol{#1}}}}
\newcommand{\mbshat}[1]{{\hat{\boldsymbol{#1}}}}
\newcommand{\mbstilde}[1]{{\tilde{\boldsymbol{#1}}}}

\newcommand{\mbsdot}[1]{{\dot {\boldsymbol{#1}}}}

\newcommand{\mbfbar}[1]{{\overline{\mbf{#1}}}}
\newcommand{\mbfhat}[1]{{\hat{\mbf{#1}}}}
\newcommand{\mbftilde}[1]{{\tilde{\mbf{#1}}}}

\newcommand{\mbfdot}[1]{{\dot{\mbf{#1}}}}

\DeclareMathAlphabet{\mathbfit}{OML}{cmm}{b}{it}

\newcommand{\vel}[3]{\leftidx{_{#1}}{\mbf v}{\IfValueTF{#2}{_{#2#3\hspace{2pt}}}{}}} 
\newcommand{\veltilde}[3]{\leftidx{_{#1}}{\mbftilde v}{\IfValueTF{#2}{_{#2#3\hspace{2pt}}}{}}} 
\newcommand{\velbar}[3]{\leftidx{_{#1}}{\mbfbar v}{\IfValueTF{#2}{_{#2#3\hspace{2pt}}}{}}} 
\newcommand{\velhat}[3]{\leftidx{_{#1}}{\mbfhat v}{\IfValueTF{#2}{_{#2#3\hspace{2pt}}}{}}} 
\newcommand{\veldot}[3]{\leftidx{_{#1}}{\mbfdot v}{\IfValueTF{#2}{_{#2#3\hspace{2pt}}}{}}} 

\newcommand{\acc}[3]{\leftidx{_{#1}}{\mbf a}{\IfValueTF{#2}{_{#2#3\hspace{2pt}}}{}}} 
\newcommand{\acctilde}[3]{\leftidx{_{#1}}{\mbftilde a}{\IfValueTF{#2}{_{#2#3\hspace{2pt}}}{}}} 
\newcommand{\accbar}[3]{\leftidx{_{#1}}{\mbfbar a}{\IfValueTF{#2}{_{#2#3\hspace{2pt}}}{}}} 

\newcommand{\rotvel}[3]{\leftidx{_{#1}}{\mbs \omega}{\IfValueTF{#2}{_{#2#3\hspace{2pt}}}{}}} 
\newcommand{\rotveltilde}[3]{\leftidx{_{#1}}{\mbstilde \omega}{\IfValueTF{#2}{_{#2#3\hspace{2pt}}}{}}} 
\newcommand{\rotvelbar}[3]{\leftidx{_{#1}}{\mbsbar \omega}{\IfValueTF{#2}{_{#2#3\hspace{2pt}}}{}}} 
\newcommand{\rotvelhat}[3]{\leftidx{_{#1}}{\mbshat \omega}{\IfValueTF{#2}{_{#2#3\hspace{2pt}}}{}}} 
\newcommand{\rotveldot}[3]{\leftidx{_{#1}}{\mbsdot \omega}{\IfValueTF{#2}{_{#2#3\hspace{2pt}}}{}}} 

\newcommand{\bx}{\mathbf{x}}
\newcommand{\bu}{\mathbf{u}}
\newcommand{\bfn}{\mathbf{f}}

\newcommand{\bw}{\mathbf{w}}
\newcommand{\bA}{\mathbf{A}}
\newcommand{\bB}{\mathbf{B}}
\newcommand{\bC}{\mathbf{C}}
\newcommand{\bI}{\mathbf{I}}
\newcommand{\set}{\mathbb}
\newcommand{\R}{\mathbb{R}}

\newcommand{\free}{\textit{free} }
\newcommand{\occupied}{\textit{occupied} }
\newcommand{\unknown}{\textit{unknown} }
\newcommand{\Vfree}{\set{V}_\text{free}}
\newcommand{\Voccupied}{\set{V}_\text{occup}}
\newcommand{\Vunknown}{\set{V}_\text{unkn}}
\newcommand{\tesdf}{h} 

\newcommand{\change}[1]{\textcolor{black}{#1}}

\title{\LARGE \bf
Control-Barrier-Aided Teleoperation with Visual-Inertial SLAM for Safe MAV Navigation in Complex Environments
}

\author{Siqi Zhou$^{1,4}$, Sotiris Papatheodorou$^{2,3,4}$, Stefan Leutenegger$^{2,3,4}$, Angela P.~Schoellig$^{1,4}$ 
\thanks{$^{1}$Learning Systems and Robotics Lab, School of Computation, Information and Technology, Technical University of Munich. E-mail addresses: \texttt{\{siqi.zhou, angela.schoellig\}@tum.de}}%
\thanks{$^{2}$Smart Robotics Lab, School of Computation, Information and Technology, Technical University of Munich. E-mail addresses: \texttt{\{sotiris.papatheodorou, stefan.leutenegger\}@tum.de}}%
\thanks{$^{3}$Smart Robotics Lab, Department of Computing, Imperial College London. E-mail addresses: \texttt{\{s.papatheodorou18, s.leutenegger\}@ic.ac.uk}}%
\thanks{$^{4}$Munich Institute of Robotics and Machine Intelligence (MIRMI).}%
\thanks{This work was supported by the Technical University of Munich, MIRMI, and the EU Horizon project DigiForest.}%
}

\begin{document}
\maketitle
\thispagestyle{empty}
\pagestyle{empty}
\sloppy

\begin{abstract}
In this paper, we consider a Micro Aerial Vehicle (MAV) system teleoperated by a non-expert and introduce a perceptive safety filter that leverages Control Barrier Functions (CBFs) in conjunction with Visual-Inertial Simultaneous Localization and Mapping (VI-SLAM) and dense 3D occupancy mapping to guarantee safe navigation in complex and unstructured environments. 
Our system relies solely on onboard IMU measurements, stereo infrared images, and depth images and autonomously corrects teleoperated inputs when they are deemed unsafe.
We define a point in 3D space as unsafe if it satisfies either of two conditions: \textit{(i)} it is occupied by an obstacle, or \textit{(ii)} it remains unmapped.
At each time step, an occupancy map of the environment is updated by the VI-SLAM by fusing the onboard measurements, and a CBF is constructed to parameterize the (un)safe region in the 3D space.
Given the CBF and state feedback from the VI-SLAM module, a safety filter computes a certified reference that best matches the teleoperation input while satisfying the safety constraint encoded by the CBF.
In contrast to existing perception-based safe control frameworks, we directly close the perception-action loop and demonstrate the full capability of safe control in combination with real-time VI-SLAM \textit{without} any external infrastructure or prior knowledge of the environment.
We verify the efficacy of the perceptive safety filter in real-time MAV experiments using exclusively onboard sensing and computation and show that the teleoperated MAV is able to safely navigate through unknown environments despite arbitrary inputs sent by the teleoperator.
\end{abstract}

\section{Introduction}

\begin{figure}
\vspace{0.5em}
    \centering
    \includegraphics[width=0.99\columnwidth]{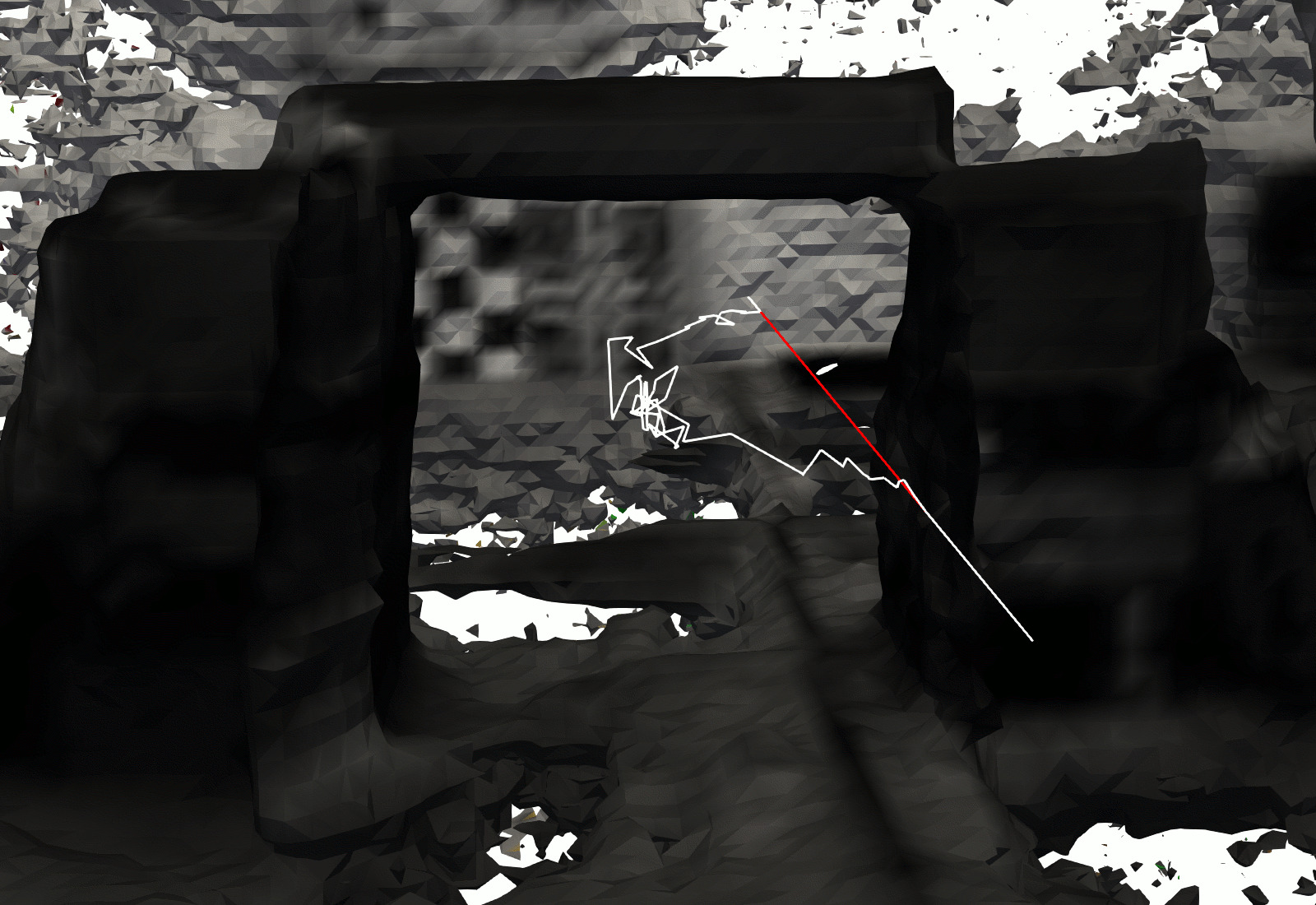}\\
    \vspace{1ex}
    \includegraphics[width=0.99\columnwidth]{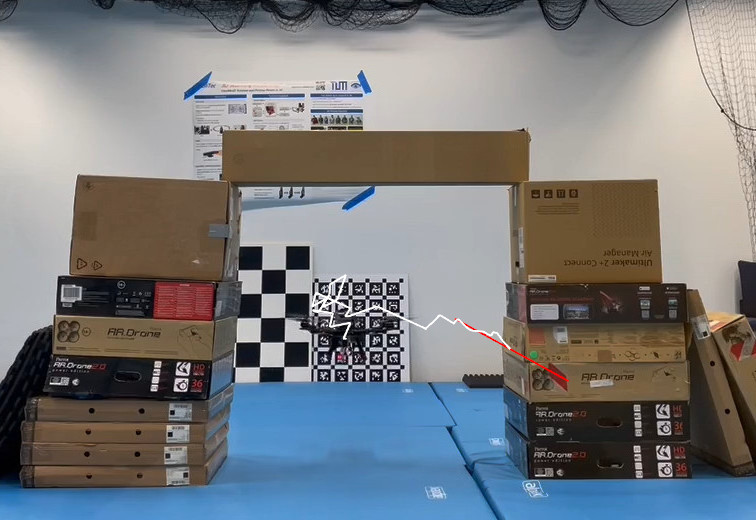}
    \caption{[Top] The operator sends a series of position references in a straight line (red). The safety filter modifies the reference (white) taking the onboard volumetric map into account. [Bottom] Photograph of the real-world MAV with the commanded (red) and filtered (white) position references overlaid.}
    \label{fig:cover}
\end{figure}

MAVs have proven to be very versatile platforms capable of aiding in a variety of tasks such as large-scale inspection, search and rescue, construction site monitoring or photography and cinematography.
While there is ongoing research on performing these kinds of tasks in an autonomous manner, direct teleoperation of the MAV by a human is at times desirable or necessary.
It is thus important to ensure safe MAV navigation not only during autonomous flight but also under direct teleoperation.

Safe autonomous flight and teleoperation differ in where the MAV reference path or trajectory originates from but have otherwise similar requirements.
In both cases, a map of the MAV's environment is required so that regions where it is safe to fly can be determined.
Creating a useful map is dependent upon the accurate localization of the MAV.
A safety-aware controller is needed to ensure no reference path causes the MAV to collide by taking the map and current MAV state estimate into account.
Finally, in order to allow deployment of the system in a wide range of environments, \change{it is} desirable for localization, mapping, and control to be performed onboard the MAV using its sensors.

In this paper, we propose a system combining onboard localization, mapping and safe control, which allows collision-free teleoperation of an MAV in complex environments.
To the best of our knowledge, this is the first system that combines a Control Barrier Function (CBF)-based safety filter with onboard, online localization and dense volumetric mapping so that no prior infrastructure or knowledge of the environment is required.
In summary, we propose the following contributions:
\begin{itemize}
    \item The generation of a Control Barrier Function (CBF) from an online dense volumetric map constructed in real-time using onboard sensors\change{, directly closing the loop between mapping and control}.
    \item The integration of VI-SLAM, dense 3D occupancy mapping, and safe control within an MAV, establishing a self-contained system that operates without the need for external infrastructure.
    \item A series of experiments conducted in both simulated environments and real-world scenarios to showcase the efficacy and robustness of our proposed system.
\end{itemize}

\section{Related Work}

We present a brief overview of the most relevant works in VI-SLAM, volumetric mapping, and safe control.

\subsection{Visual-Inertial Odometry and SLAM}

Visual-Inertial Odometry (VIO) is the process of estimating a robot's state given camera images and Inertial Measurement Unit (IMU) readings.
The IMU measurements are especially important in the case of MAVs which can exhibit fast movement that may cause a visual-only odometry system to lose track of the MAV's state.
Visual-Inertial SLAM expands on VIO by also performing loop closure upon place recognition and therefore keeping consistent estimates of trajectory and map.

One of the first approaches to combine visual with inertial measurements was MSCKF \cite{msckf}, a filtering-based approach for adding geometric constraints from visual observations into an inertial navigation algorithm.
There have been several extensions to this method since (e.g., OpenVINS~\cite{openvins}) and other notable filter-based approaches (e.g., ROVIO~\cite{rovio}).
It is extended in maplab \cite{maplab} to perform localization against large-scale sparse maps.

Non-linear optimization based approaches such as BASALT \cite{basalt}, VINS-Fusion \cite{vinsfusion} and ORB-SLAM3 \cite{orbslam3} are also popular due to their improved accuracy.
In Kimera \cite{kimera}, a sparse VI-SLAM system is combined with dense mapping using meshes.
OKVIS \cite{okvis} is a lightweight optimization-based VIO method that is extended in OKVIS2 \cite{okvis2} to include loop closures and semantic segmentation for the filtering of dynamic objects. Due to its state-of-the-art accuracy and runtime performance, we use it in our approach.

\subsection{Volumetric Mapping}

Volumetric mapping refers to the creation of a dense 3D map of an environment which can then be used for SLAM or planning.
In the seminal work of KinectFusion~\cite{kinect_fusion}, the mapped space is represented as a discretized Truncated Signed Distance Field (TSDF) stored in a fixed-size grid.
Newer approaches have improved the scalability of the map by using voxel hashing \cite{infinitam} or octrees \cite{supereight} instead of a grid.
TSDF-based maps don't explicitly represent \free space and thus cannot be used directly for safe navigation.
Voxblox~\cite{voxblox} solves this issue by incrementally creating a Euclidean Signed Distance Field (ESDF) from the TSDF map to allow safe onboard motion planning;
\change{however, it does not directly close the loop between mapping and control.}

Another approach is to perform occupancy mapping, which entails storing the occupancy probability of each point in the mapped space.
This has the benefit that \textit{occupied}, \textit{free} and \textit{unknown} regions of the environment can be explicitly represented, allowing safe path planning.
OctoMap~\cite{octomap} is a popular occupancy mapping framework that has been the basis of other methods such as UFOMap~\cite{ufomap}, which extends it to explicitly store \unknown space.
FIESTA~\cite{fiesta} incrementally creates an ESDF map suitable for MAV motion planning from an occupancy map.
Another octree-based occupancy mapping method is supereight~\cite{supereight} and its extension \cite{supereight2}, which introduces propagation of occupancy values through all levels of the octree for efficient path planning.

\subsection{Safe Control}
Control theory has been widely used to provide desired safety guarantees for dynamical systems. In recent years, there has been an increasing number of work on learning-based methods to further address the challenge of guaranteeing the safety of the system in the presence of uncertainties~\cite{brunke2022safe}. Examples include but are not limited to Gaussian Process Model Predictive Control (GP-MPC)~\cite{ostafew2016learning}, GP-based robust control~\cite{berkenkamp2015safe}, neural adaptive control~\cite{joshi2019deep}, and learning-based Control Barrier certification~\cite{taylor2020learning}. While classical control and more advanced learning-based control methods have proven effective, most of them rely on having sufficiently accurate state feedback and a clear understanding of the environment~\cite{brunke2022safe}. In practical applications, robot systems are subject to noisy state estimation and incomplete knowledge about their surrounding environments. In these cases, they are required to distill safety constraints in real-time from high-dimensional sensory inputs and comply with safety constraints inferred from perception. 

There exist works targeted to bridge the gap between perception and safe control. For instance, a robust CBF formulation was introduced in \cite{dean2021guaranteeing} to account for perception errors, a differentiable barrier network was proposed in \cite{xiao2023barriernet} to incorporate a CBF quadratic program (QP) in an end-to-end learning pipeline, and a perceptive safe locomotion approach was presented in \cite{grandia2023perceptive} to infer steppable planes and safe actions based exteroceptive sensors. In this work, we similarly aim to bridge the perception-action gap with a particular emphasis on demonstrating the capabilities of a safe control system that incorporates VI-SLAM and dense 3D mapping in the closed loop for safe teleoperated MAV navigation in unstructured and cluttered environments.

\section{Problem Statement}

We consider an MAV that tracks reference positions produced by a human operator.
Our goal is to ensure the MAV safely navigates cluttered environments and avoids collisions even when the operator reference positions would cause it to collide. Additionally, the MAV should be prevented from entering unmapped regions.
To achieve these, we use the MAV's onboard sensors to localize within the environment, create a map of safely navigable regions, and use a safety filter to modify the operator reference positions as required to avoid constraint violations.
We assume that the MAV is equipped with an IMU, a stereo camera and a depth sensor.

\subsection{Environment Model}

The static environment is modelled as a bounded volume $\set{V} \subset \set{R}^3$ with each point $\mbf{x} \in \set{V}$ having an associated occupancy probability $P_o(\mbf{x})$.
We define the \free, \unknown and \occupied sets as $\Vfree = \{\mbf{x} \in \set{V}, \ P_o(\mbf{x}) < 0.5\}$,
$\Vunknown = \{\mbf{x} \in \set{V}, \ P_o(\mbf{x}) = 0.5\}$ and
$\Voccupied = \{\mbf{x} \in \set{V}, \ P_o(\mbf{x}) > 0.5\}$.

\subsection{MAV Control System}
The MAV control system computes desired attitude commands based on position references. We assume that the closed-loop MAV system can be locally approximated by a discrete-time linear model:
\begin{equation}
    \mbf{x}_{k+1} = \bfn(\mbf{x}_k,\bu_k,\bw_k) \approx \bA_k\mbf{x}_k + \bB_k\bu_k + \bw_k,
\end{equation}
where $k\in\set{Z}_{\ge 0}$ denotes discrete-time index, $\mbf{x}\in\set{V}\subset \R^3$ is the position of the MAV system, $\bu\in\set{U}\subset\R^3$ is the position reference input, $\bw_k$ is random noise, $\bfn: \set{V}\times \set{U}\mapsto \set{V}$ is a Lipschitz continuous function, and $(\bA,\bB)$ are matrices of consistent dimensions. 

In this work, we assume that the noise in the system is bounded as follows: $||\bw_k||\le \epsilon$ with $\epsilon\in\R_{\ge 0}$, where $||\cdot||$ denotes the Euclidean norm. When the control system has perfect tracking, the system matrices $\bA_k$ and $\bB_k$ would be a zero matrix and an identity matrix, respectively. When perfect tracking is not achieved, one may learn the system matrices $(\bA_k, \bB_k)$ based on data collected online to account for the dynamics of the underlying closed-loop system.

\section{Methodology}

\begin{figure}
    \centering
    \includegraphics[width=\columnwidth]{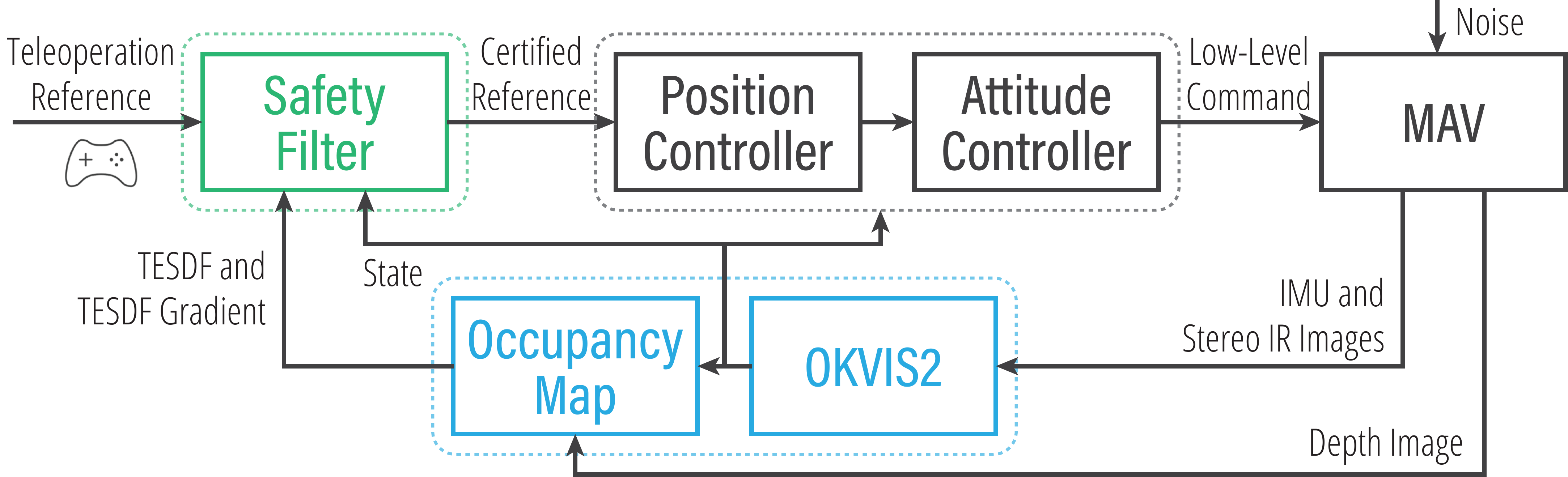}
    \caption{A block diagram of the MAV system. The system consists of three main modules: \textit{(i)} a baseline controller for reference tracking, \textit{(ii)} a perception module (blue) that keeps an updated dense 3D map of the environment and provides estimates of the MAV state in real-time, and \textit{(iii)} a safety filter (green) that adjusts the teleoperated inputs based on perceptive information when they are deemed unsafe. Our goal is to design a perceptive safety filter to guarantee the safety of the teleoperated MAV system in complex and unstructured environments. 
    }
    \label{fig:block_diagram}
\end{figure}

Our approach combines VI-SLAM, dense 3D mapping, and a safety filter, all running online and onboard the MAV.
The OKVIS2 \cite{okvis2} VI-SLAM system receives grayscale stereo image pairs and IMU measurements and produces MAV state estimates.
Depth images and corresponding pose estimates are used by supereight 2 \cite{supereight2} to create a dense 3D occupancy map.
The control system of the MAV consists of a position controller and an attitude controller, which together serve as a tracking controller that converts desired position signals to lower-level commands.
The lower-level commands are then realized by the MAV autopilot.
The safety filter layer utilizes the map generated by the perception system and minimally adjusts the reference sent to the controller to ensure collision avoidance with obstacles and remaining within mapped regions.
A block diagram of the overall system architecture can be seen in Fig.~\ref{fig:block_diagram}.

\subsection{Visual-Inertial SLAM}

We use the OKVIS2 \cite{okvis2}, a state-of-the-art, sparse, optimization-based VI-SLAM system for MAV state estimation.
OKVIS2 consists of a frontend and a real-time estimator that processes multiple images and IMU messages
synchronously whenever a new frame arrives.
The front end performs keypoint matching, stereo triangulation and place recognition which triggers loop closures if successful.
The real-time estimator optimizes the current factor graph and marginalizes old observations.
OKVIS2 also performs optimization of the full factor graph on loop closures asynchronously and then synchronizes the result with the real-time factor graph.
We further utilize the real-time IMU propagation capabilities of OKVIS2 to obtain the most up-to-date state estimates at IMU rate, numerically integrating from the newest optimized state.

\subsection{Dense Mapping}

We use a modified version of the multi-resolution mapping pipeline from \cite{supereight2} to create a dense volumetric occupancy map of the environment.
Depth measurements are integrated in a probabilistic manner to account for sensor noise, and \free space is explicitly mapped to allow safe MAV navigation.
It is important to note that for collision-free MAV movement, only \free regions of the map must be considered safe since \unknown regions may contain yet unmapped obstacles.

\begin{figure}[hbt]
    \centering
    \includegraphics[width=0.99\columnwidth]{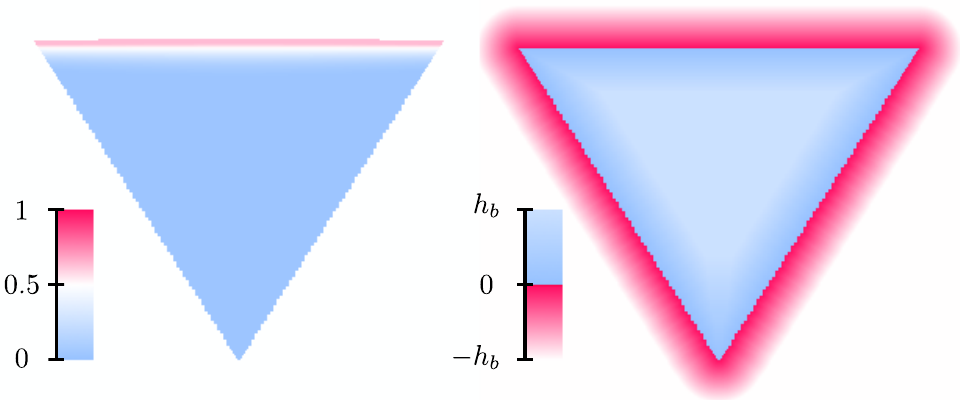}
    \caption{[Left] Occupancy field slice after integration of a single depth image with \textit{free}, \textit{occupied}, and \textit{unknown} space shown in shades of blue, red and white, respectively. [Right] Corresponding TESDF slice with positive (safe) and negative (unsafe) values shown in shades of blue and red, respectively. Notice the occupancy field discontinuity between \textit{free} and \textit{unknown} space compared to the smooth transition of the TESDF.}
    \label{fig:occupancy_tesdf}
\end{figure}

In order to use the occupancy map in a CBF, a differentiable field must be extracted from it.
The occupancy field is discontinuous at the boundaries between \free and \unknown space, as shown in Fig.~\ref{fig:occupancy_tesdf} [Left], making it unsuitable for use with a CBF.
A space representation more suitable for use with a CBF is the Truncated Euclidean Signed Distance Function (TESDF) $\tesdf(\mbf{x}): \set{V} \rightarrow \set{R}$ which is differentiable over $\set{V}$ and defined as
\begin{equation}
    \tesdf(\mbf{x}) = 
    \begin{cases}
        \ \ \,\min\left( d(\mbf{x}, \ \partial \Vfree), \ \tesdf_b \right), &\text{if } \mbf{x} \in \Vfree, \\
        -\min\left( d(\mbf{x}, \ \partial \Vfree), \ \tesdf_b \right), &\text{otherwise},
    \end{cases}
\end{equation}
where $\tesdf_b \in \set{R}^+$ is the truncation bound, $\partial V$ denotes the boundary of set $\set{V}$ and $d(\mbf{x}, \set{V})$ the Euclidean distance of point $\mbf{x}$ from set $\set{V}$.
This results in the TESDF having positive values in the interior of the safe set (i.e.,\ \free space), negative values outside the safe set (i.e.,\ \occupied or \unknown space), and 0 at the safe set boundary.
The truncation bound $\tesdf_b$ is used to limit the volume affecting the TESDF of a given point, allowing more efficient computation of TESDF values in a volumetric map.

We extend the mapping framework to allow querying $\tesdf(\mbf{x})$ and $\nabla \tesdf(\mbf{x})$ for any $\mbf{x} \in \set{V}$.
The queried TESDF values and gradients are computed online from the latest occupancy map data to ensure safe navigation.
A TESDF slice through the map and the corresponding occupancy slice are shown in Fig.~\ref{fig:occupancy_tesdf}.

\subsection{Safety Filter Design}

We employ a discrete-time CBF formulation in the design of our safety filter. To facilitate our discussion, we first introduce the case where $\bw_k=\mathbf{0}$ and then extend the framework to account for noise in the system.

A discrete-time CFB safety filter can be formulated as an optimization problem~\cite{zeng2021safety}:
\begin{subequations}
    \begin{align}
        \min_\bu &\hspace{1em}||\bu - \bu_\text{teleop}||^2\\
       \text{subject to} &\hspace{1em} \Delta \tesdf(\mbf{x}_k, \bu) \ge \alpha\: \tesdf(\mbf{x}_k),\label{subeqn:cbf-condition}
    \end{align}
    \label{eqn:safety-filter}%
\end{subequations}
where $\Delta \tesdf(\mbf{x}_k, \bu) = \tesdf\big(f(\mbf{x}_k, \bu)\big) - \tesdf(\mbf{x}_k)$ is the change in the value of the TESDF after an input $\bu$ is applied, and $\bu_\text{teleop}$ is the teleoperation input. The term $\alpha$ is defined as follows: 
\begin{equation}
    \alpha = \begin{cases}
        -p_1, &\hspace{1em}\text{ if } \tesdf(\mbf{x}_k) \ge 0,\\
       \hspace{0.75em} p_2, &\hspace{1em}\text{ otherwise},
    \end{cases}
\end{equation}
where $p_1$ and $p_2$ are positive constants. The safety filter in~\eqref{eqn:safety-filter} aims to find an input $\bu$ that is close to the teleoperation input~$\bu_\text{teleop}$ while guaranteeing the safety constraint represented by the CBF condition is satisfied.

Intuitively, the CBF condition~\eqref{subeqn:cbf-condition} is a scalar condition that guarantees the positive invariance of a set (i.e., if the system starts in the set, it will remain in the set). When the system is in the safe set, the condition has two implications: \textit{(i)} at the safety boundary, the change in $\tesdf$ is lower bounded by 0, which implies that the system either remains on the boundary or moves towards the interior of the safe set, and \textit{(ii)} in the interior of the safe set, the change in $\tesdf$ is lower bounded by a negative number which quantifies how fast the system is allowed to approach the safety boundary. When the system is outside of the safe set, the change in $\tesdf$ is required to be positive, which is intended to move the system closer to the safe set and ultimately return to the safe set. 

The CBF constraint in the optimization problem in~\eqref{eqn:safety-filter} is generally nonlinear in the decision variable $\bu$ and is not trivial to solve. To simplify the problem, we use a first-order approximation of the CBF constraint, which allows us to formulate the safety filter optimization problem as a quadratic program (QP) that can be solved efficiently online. The first-order approximation of the CBF constraint can be written as follows:
\begin{equation}
     \bC \:\bu \ge c_1,
     \label{eqn:linearized-cbf}
\end{equation}
where $\bC \mathord{=} \nabla^\intercal h(\bx_k)\:\bB_k$ and $c_1 \mathord{=}  \nabla^\intercal h(\bx_k)\: \left(\bI - \bA_k\right)\bx_k +\alpha h(\bx_k)$.

When the system dynamics are noisy, one can robustly account for the noise in the design of the CBF condition to guarantee safety in the presence of uncertainties~\cite{dean2021guaranteeing}.
We introduce a robust CBF condition in our safety filter design. By introducing the term $\bw$ to $\Delta h$ and following the same linearization procedure, we obtain the following robustified condition:
\begin{equation}
         \bC \:\bu \ge c_1 + c_2,
\end{equation}
where $c_2 = ||\nabla h(\bx_k)||\epsilon$. This condition increases the lower bound on $\Delta h$ by a positive number that is proportional to the level of noise in the system. Intuitively, the term $c_2$ characterizes the worst-case uncertainty in $\Delta h$ due to system noise. The overall CBF safety filter is
\begin{subequations}
    \begin{align}
        \min_\bu &\hspace{1em}||\bu - \bu_\text{teleop}||^2\\
       \text{subject to} &\hspace{1em} \bC \:\bu \ge c_1 + c_2,
    \end{align}
    \label{eqn:safety-filter-robust}%
\end{subequations}
which has a standard QP form.

To account for the spatial extent of the MAV, one can generally offset the TESDF such that $h \ge \underline{d}$ is guaranteed, where $\underline{d}$ characterizes the dimension of the MAV or apply the CBF condition to a set of sample points on the collision envelope of the MAV. The latter could result in less conservative behaviour but incurs additional computation costs.

\section{Evaluation}

In order to evaluate our method, we performed several experiments both in simulation and on a real MAV. The MAV is teleoperated by a human operator. The teleoperation inputs consist of position and yaw references that drive the MAV from one position to another in the environment. We apply the proposed CBF safety filter to the raw position inputs provided by the teleoperator and generate certified position references that ensure \textit{(i)} the MAV does not collide with any obstacles in the environment and \textit{(ii)} the MAV does not enter areas that are not yet mapped. The yaw reference provides an additional degree of freedom for mapping the unknown environment but does not directly affect the safety of the MAV system. We therefore do not include the yaw reference in our safety filter design.

In both simulated and real-world experiments, OKVIS2 is configured to produce state estimates at 60 Hz, which are used by the position controller also operating at 60 Hz. \change{The safety filter is implemented using qpOASES~\cite{ferreau2014qpoases} and is applied in real-time at 6 Hz.} A map of the environment with a resolution of 0.05 m is updated using $640 \times 480$ depth images, and the truncation bound $h_b$ for the TESDF is 0.5~m. In the CBF certification optimization problem, we use a slope of $p_1 = 0.45$ for the safe set to bound how fast the MAV is allowed to approach the safety boundary and a slope of $p_2 = 1\times 10^{-3}$ for the unsafe region to define how quickly the MAV is required to converge to the safe region if it starts outside of the safe set. The noise parameter $\epsilon$ is set to 0.1~m in the simulation and 0.2~m in the experiment.

\subsection{Simulation Results}

\begin{figure}[t]
    \centering
    \includegraphics[width=\columnwidth]{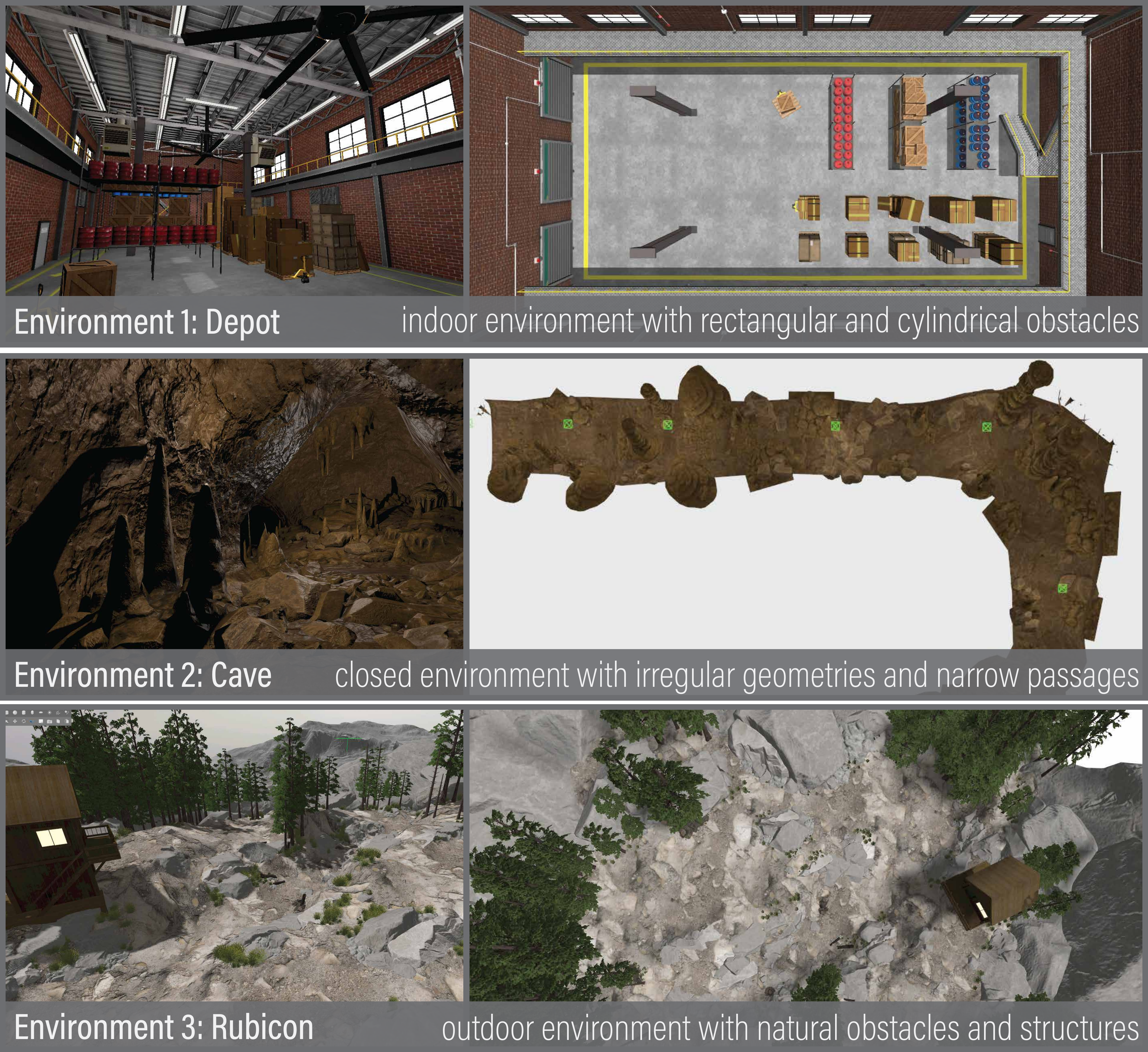}
    \caption{Visualizations of the simulation environments used in our evaluation, each with \change{different obstacle types} and level of clutteredness~\cite{gazebo}.}
    \label{fig:sim-environments}
\end{figure}

We use the Gazebo \cite{gazebo} simulator with the PX4 autopilot \cite{px4} simulated in software.
This results in a simulation with realistic MAV dynamics and sensors and a control interface that is exactly the same as the real-world MAV.
The MAV model is based on the RMF-Owl \cite{rmfowl}, a $0.38 \times 0.38 \times 0.24$~m quadcopter, that is equipped with an Intel RealSense D455 RGB-D camera.
We evaluate our approach in three different simulated environments \change{with different obstacle types and levels of clutteredness} (as depicted in Fig.~\ref{fig:sim-environments}).

\begin{figure}[t]
    \centering
    \includegraphics[width=\columnwidth]{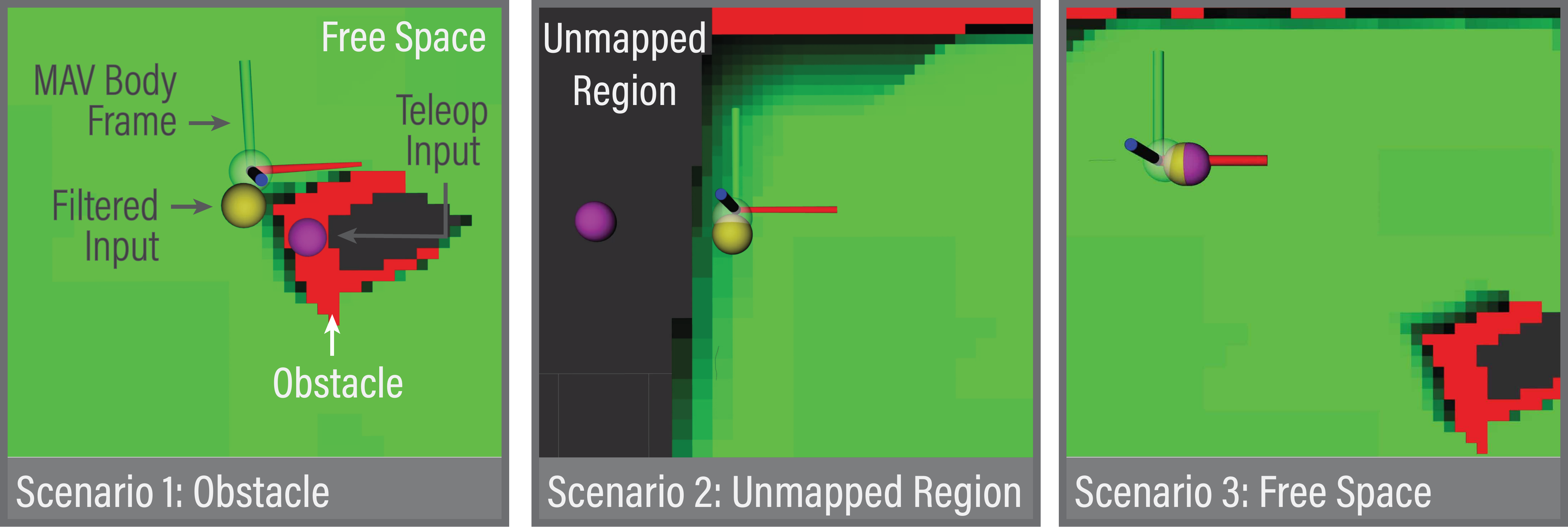}
    \caption{Example cases from the evaluation in the Depot environment. In these plots, free space is indicated in green, obstacles are marked in red, and unknown area is shown in dark grey. The teleoperation input and the filtered input are indicated by purple and yellow spheres, respectively. The MAV position is drawn as a transparent sphere with its body frame attached. When the teleoperation input is unsafe---colliding with an obstacle [Left] or leading into the unmapped region [Middle], the safety filter finds the closest reference that does not violate safety constraints. In the interior of the safe region [Right], the filtered reference coincides with the teleoperation input.}
    \label{fig:simulation-filter-example}
\end{figure}

\begin{figure}[ht]
    \centering
    \includegraphics[width=\columnwidth, trim={0.19cm, 0cm, 0.2cm, 0cm}, clip]{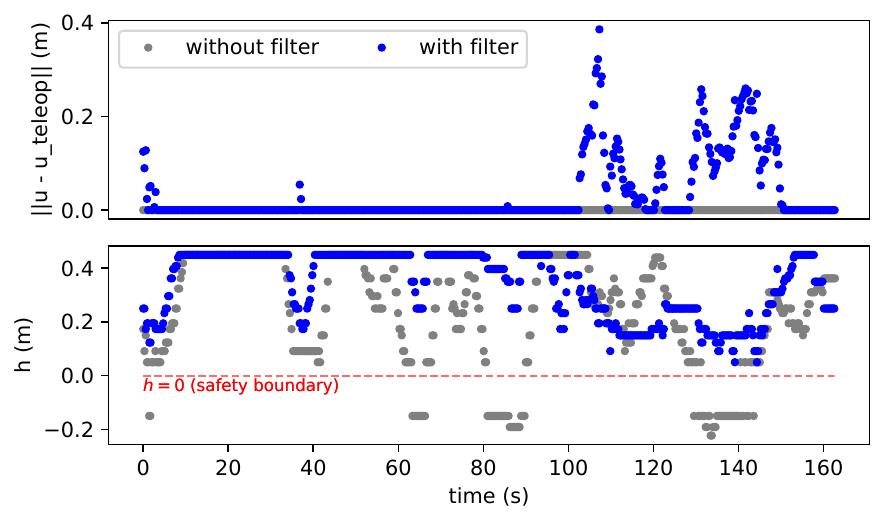}
    \caption{Example trajectories illustrating the input correction applied by the safety filter and the corresponding CBF as the MAV is teleoperated in the Depot environment. The safety filter applies larger input corrections when the MAV is close to the safety boundary ($h=0$) and applies approximately zero corrections when the MAV is in the free space (high $h$ values). Without the safety filter, the teleoperated reference led to collisions at multiple time instances, while the filtered signals kept the MAV inside of the safe set.}
    \label{fig:warehouse-traj}
\end{figure}

Figure~\ref{fig:simulation-filter-example} illustrates the application of our proposed safety filter design. In the left and middle panels of the figure, we see that the raw references (depicted as magenta spheres) are directed toward either obstacles or unmapped regions. The filtered inputs (shown as yellow spheres) effectively constrain the robot (represented by the transparent sphere with the body frame attached) to remain within the safe region. Over the free space, as seen in the right panel, the safety filter does not modify the teleoperation input, and the MAV closely follows the raw teleoperation input. Fig.~\ref{fig:warehouse-traj} shows a set of quantitative results corresponding to a test trial in the first simulation environment. When the MAV is close to the safety boundary $(h = 0)$, the safety filter applies a larger correction to move the robot away from the boundary to prevent possible constraint violations and maintains approximately zero input correction over safe regions (with high $h$ values). Over the course of this trajectory, the raw teleportation reference led to collisions at~83 discrete time instances, while the filtered reference led to zero constraint violation. 

\begin{figure*}[tb]
    \centering
    \begin{tabular}[t]{c @{\hspace{2pt}} c @{\hspace{2pt}} c @{\hspace{2pt}} c}
          \includegraphics[width=0.241\textwidth, trim={2cm, 0cm, 2cm, 0cm}, clip]{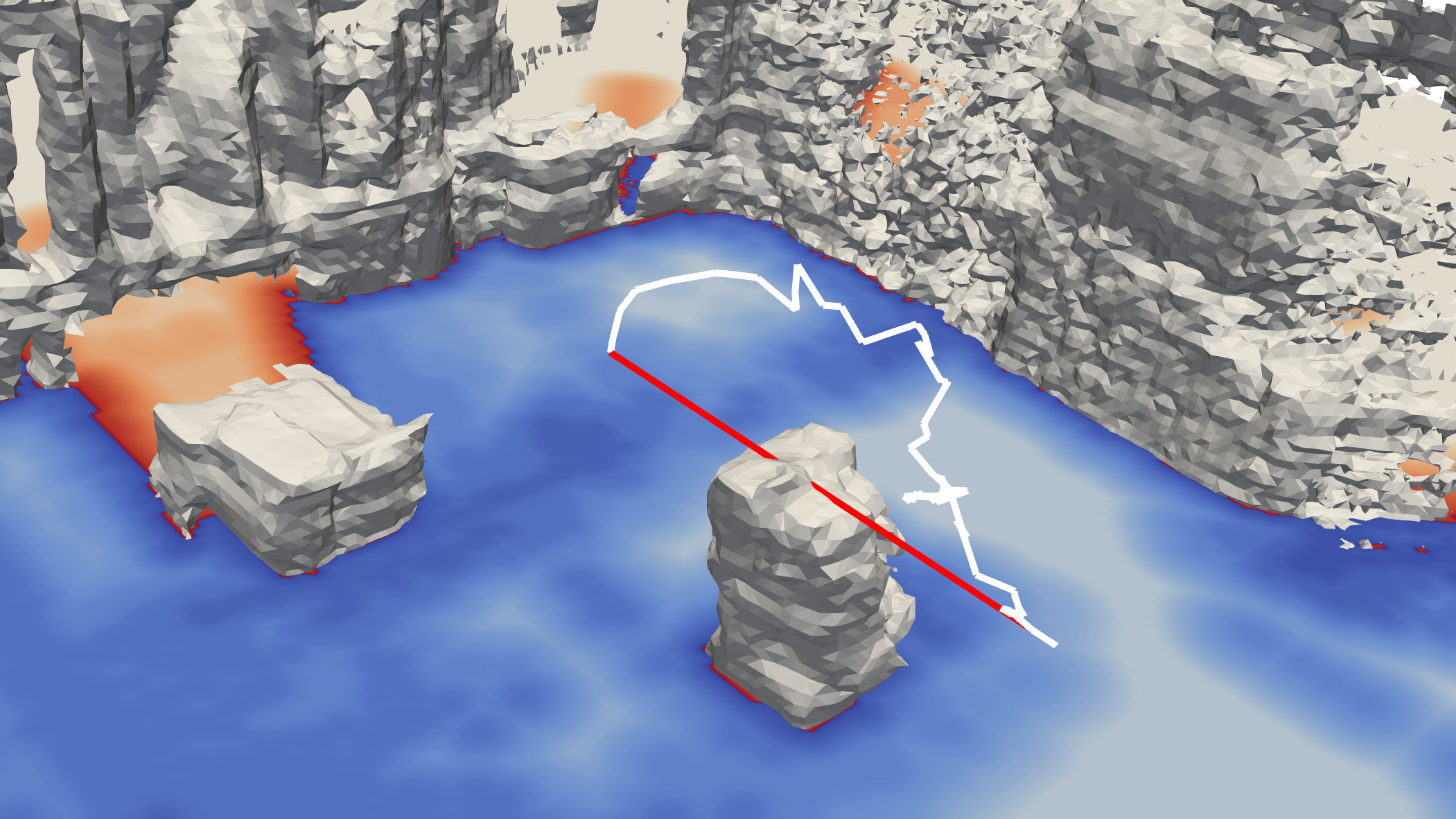}
        & \includegraphics[width=0.241\textwidth, trim={2cm, 0cm, 2cm, 0cm}, clip]{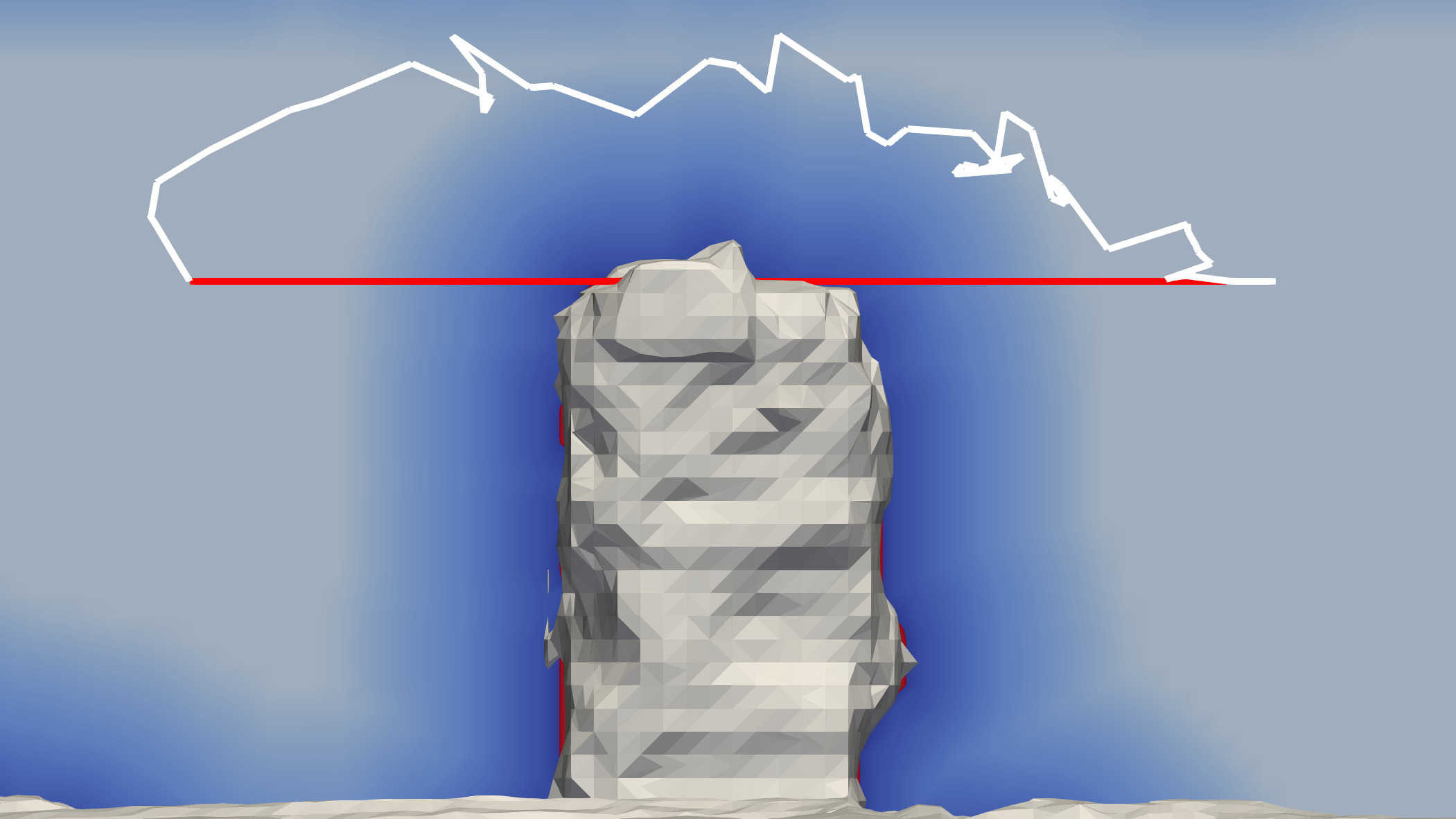}
        & \includegraphics[width=0.241\textwidth, trim={2cm, 0cm, 2cm, 0cm}, clip]{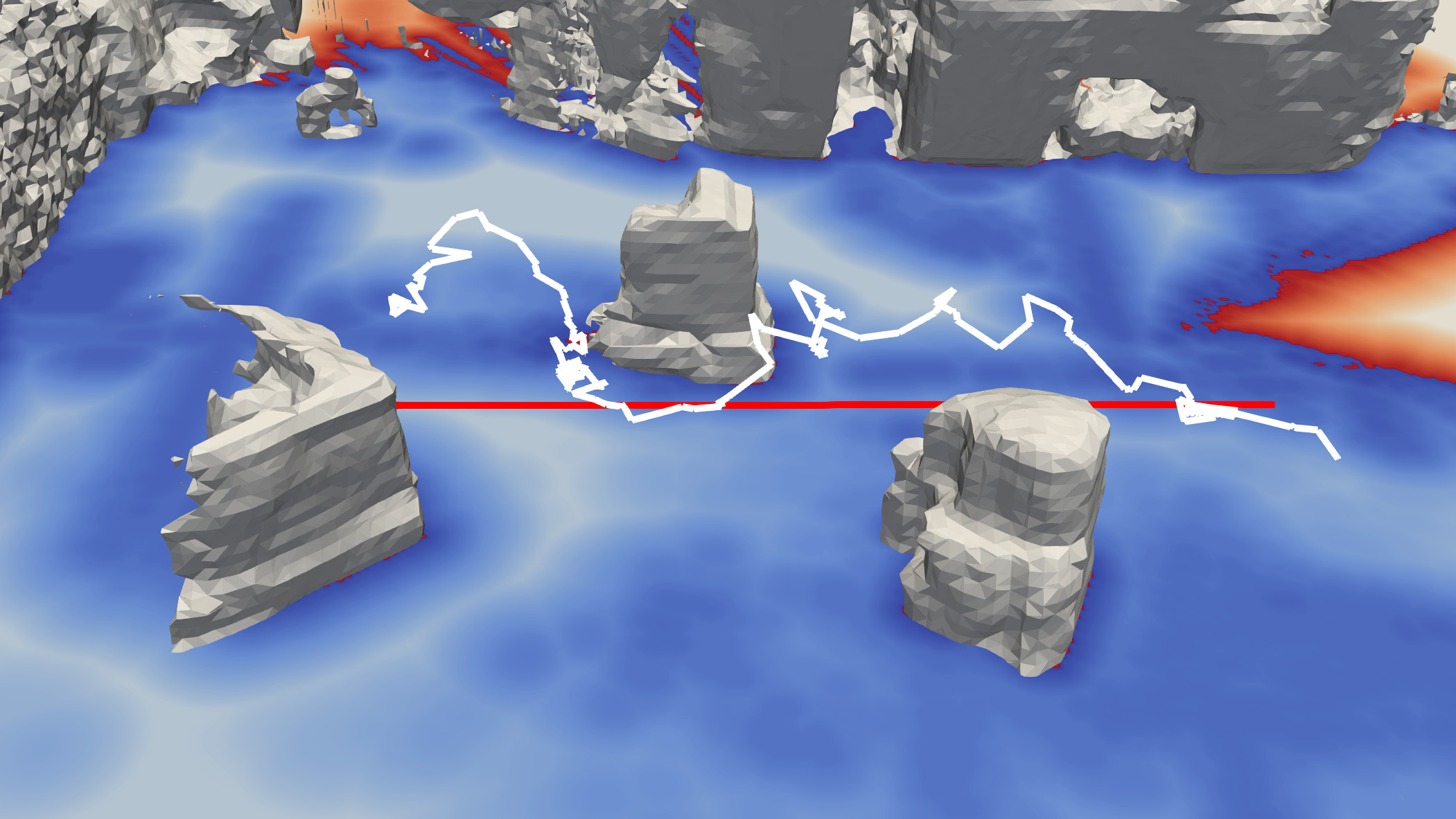}
        & \includegraphics[width=0.241\textwidth, trim={2cm, 0cm, 2cm, 0cm}, clip]{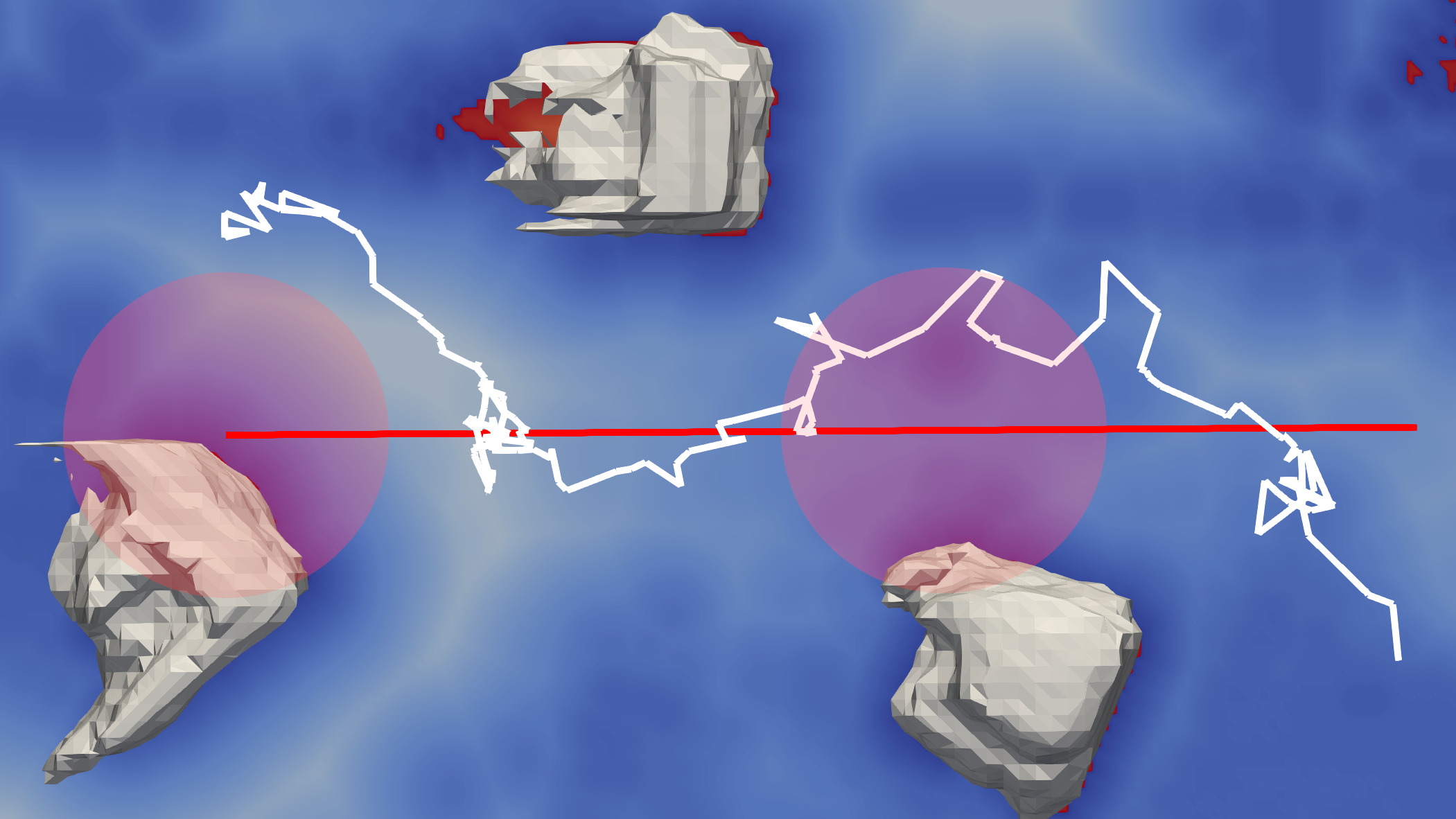}\\
        \vspace{4pt}
        (a) & (b) & (c) & (d)\\
          \includegraphics[width=0.241\textwidth, trim={2cm, 0cm, 2cm, 0cm}, clip]{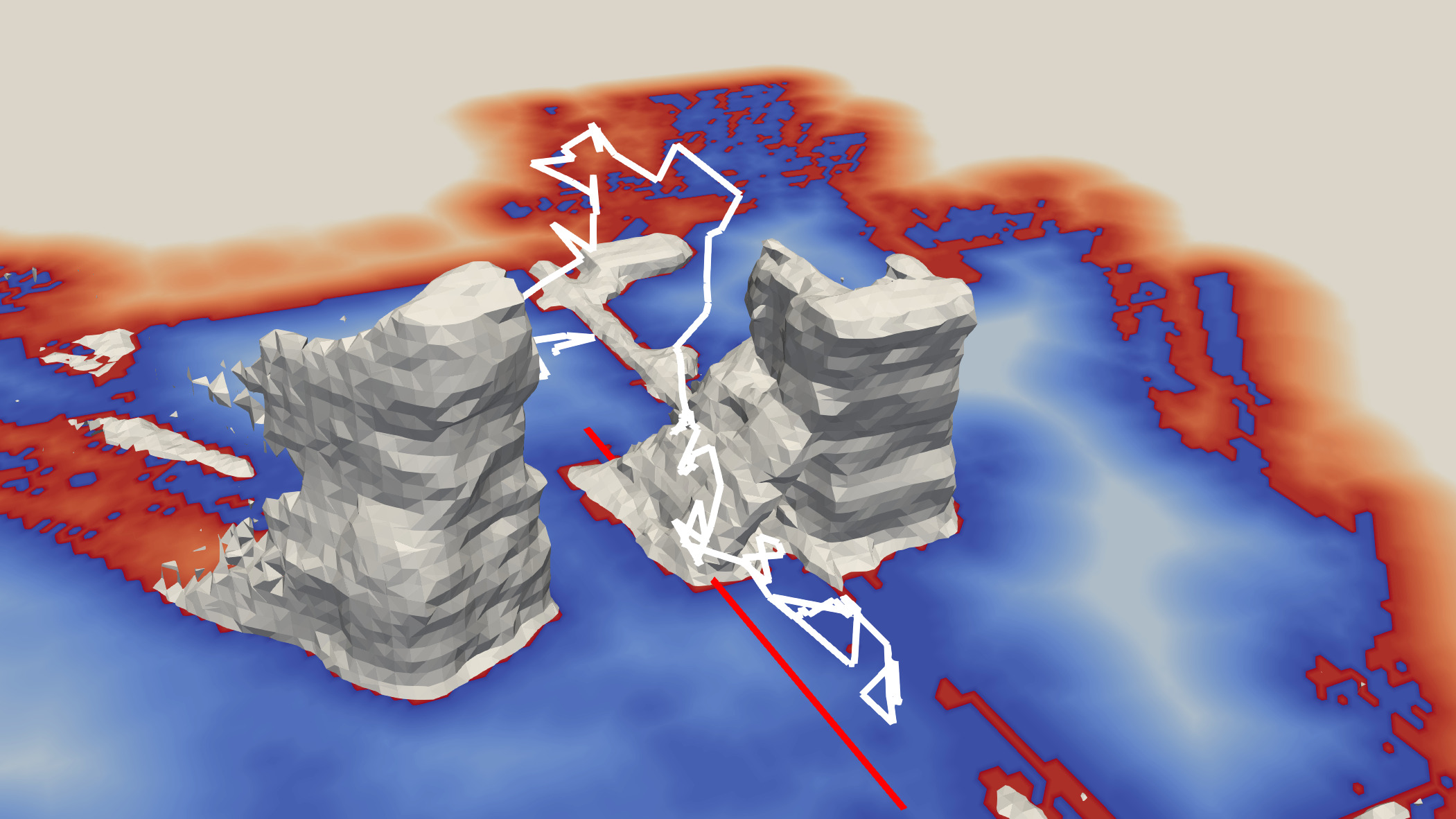}
        & \includegraphics[width=0.241\textwidth, trim={2cm, 0cm, 2cm, 0cm}, clip]{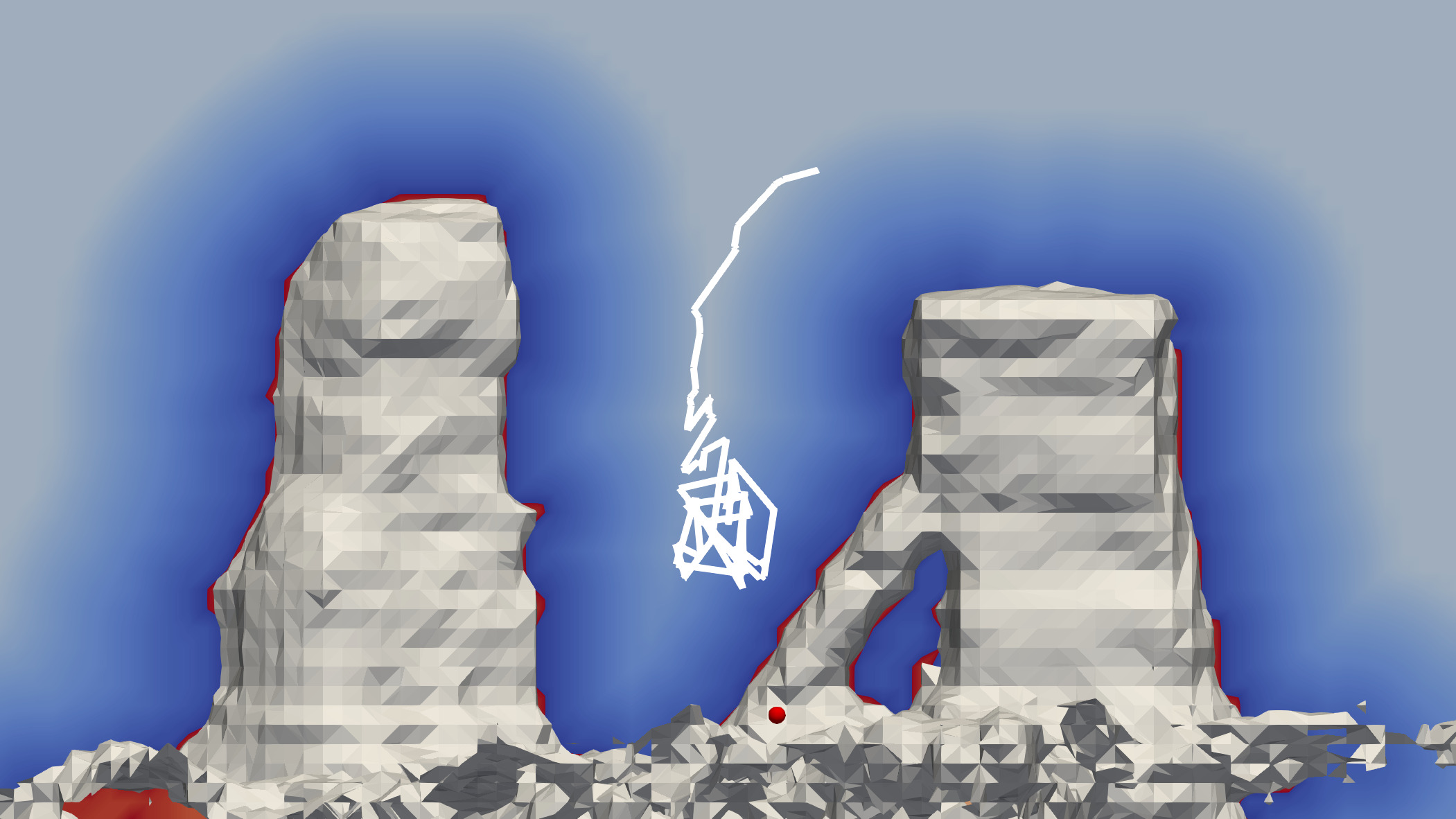}
        & \includegraphics[width=0.241\textwidth, trim={2cm, 0cm, 2cm, 0cm}, clip]{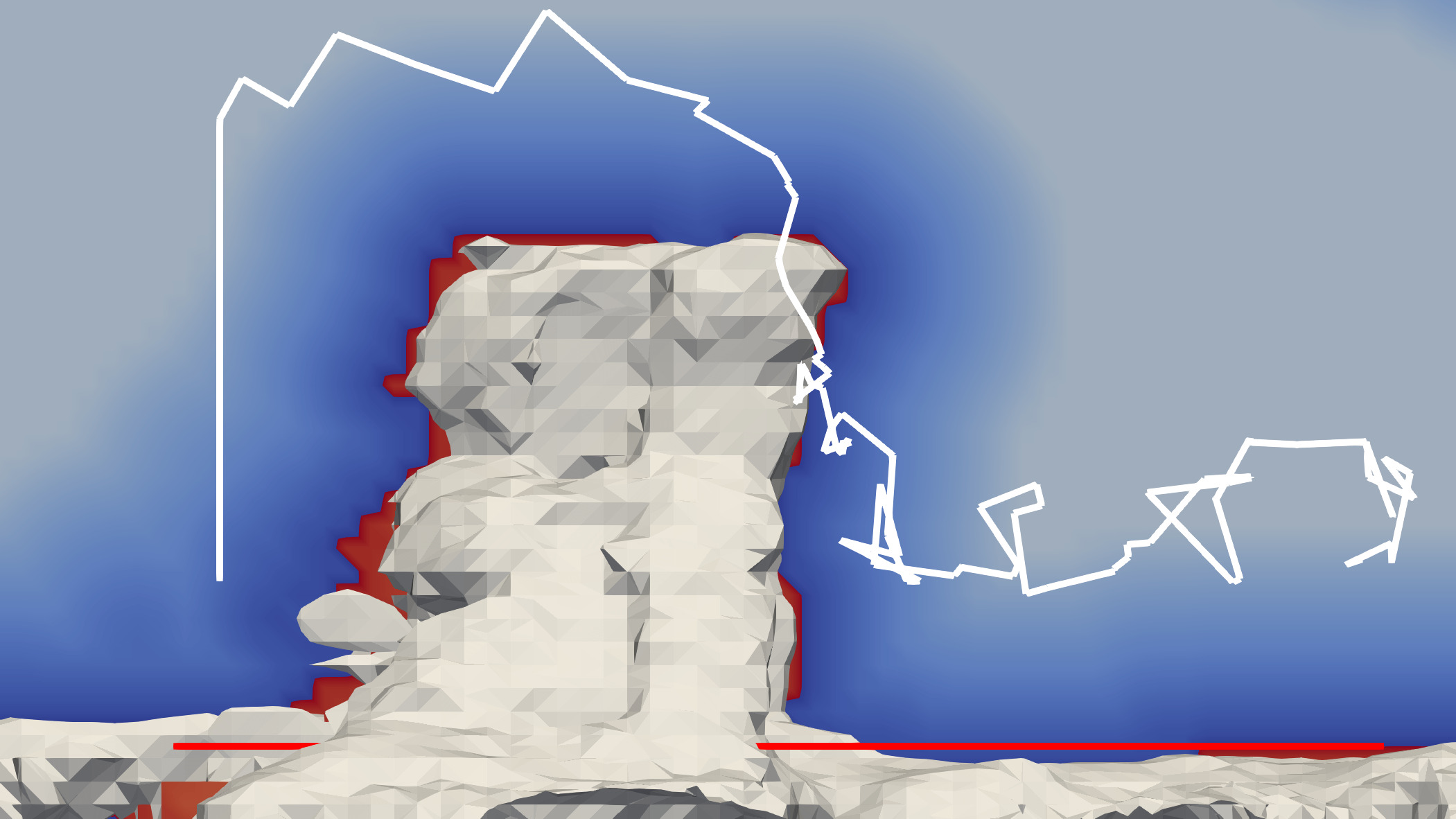}
        & \includegraphics[width=0.241\textwidth, trim={2cm, 0cm, 2cm, 0cm}, clip]{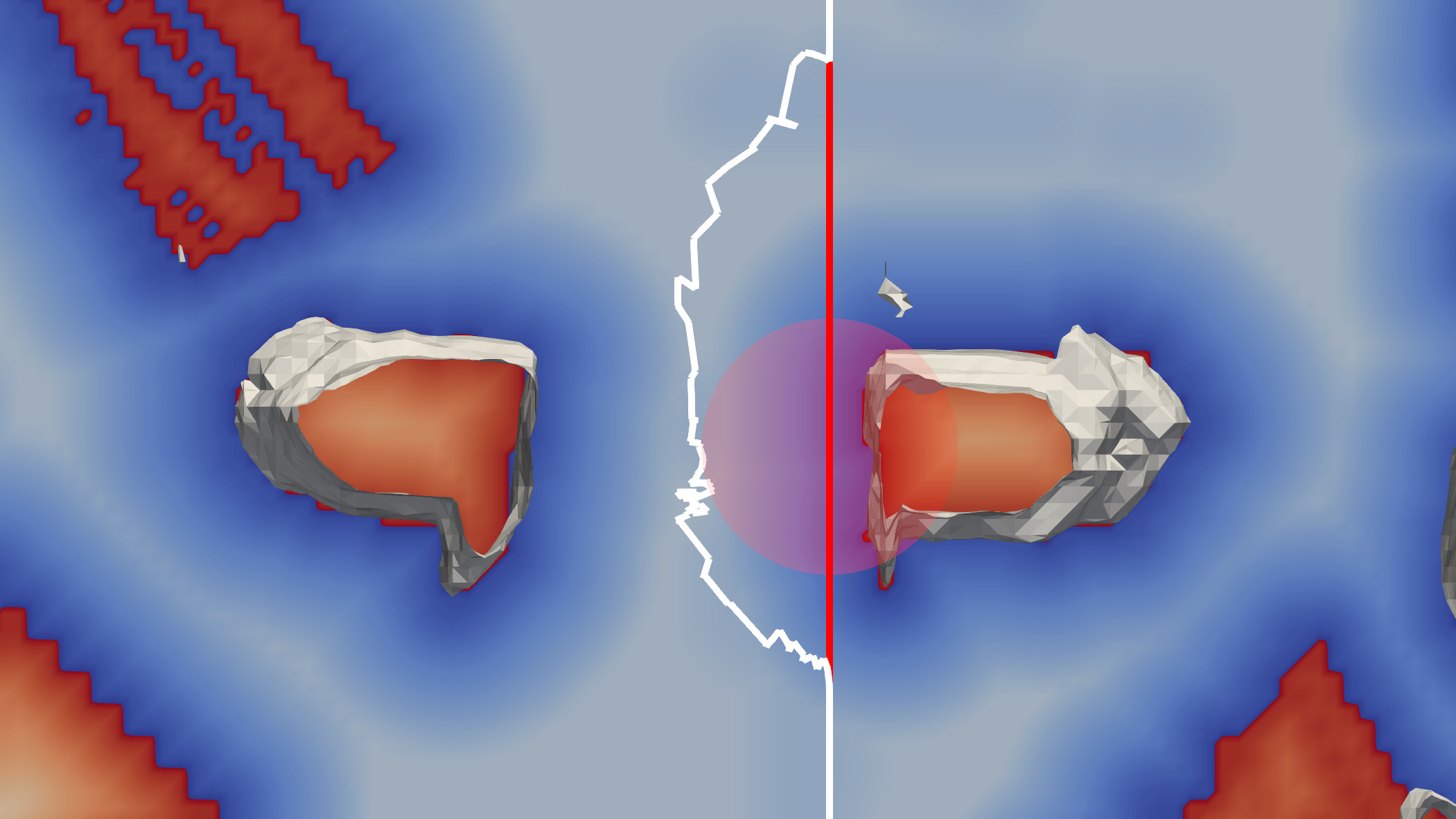}\\
        (e) & (f) & (g) & (h)
    \end{tabular}
    \caption{Visualizations of the commanded (red) and filtered (white) reference position trajectories in real-world experiments. Positive (safe) TESDF values are shown in shades of blue, while negative (unsafe) values are in shades of red, as in Fig.~\ref{fig:occupancy_tesdf} [Right]. Meshes extracted from the occupancy map are shown in gray, with some of the walls removed for clarity. Red circles with the MAV dimensions are shown to highlight some of the collisions that would have happened had the original reference been followed.
    (a), (b): Perspective and side view of an experiment where the commanded reference position passed through an obstacle.
    (c), (d): Perspective and top-down view of an experiment where the commanded reference position was too close to obstacles.
    (e), (f), (g): Perspective, front and side view of an experiment where the commanded reference position was too close to obstacles and the ground.
    (h): Top-down view of the experiment from Fig.~\ref{fig:cover} where the commanded reference position was too close to obstacles.}
    \label{fig:experiment}
\end{figure*}

Table~\ref{tab:sim-result-summary} presents a summary of our evaluation across three distinct simulation environments (Fig.~\ref{fig:sim-environments}), each comprised of three independent teleoperated trials. The trials corresponding to the teleoperated cases without the safety filter exhibited varying degrees of constraint violations, as indicated by the $h_\text{min}$ values. Notably, The safety filter responded by applying larger input corrections, as quantified by $||\delta \bar{u}||$, in trials with more substantial constraint violations. Through the application of our safety filter, we have consistently achieved effective safety assurance for the teleoperated MAV system. The average adjustment in input introduced by the safety filter ranged from 0.03 m to 1.00 m. The $h_\text{min}$ values spanned from -0.50 m to -0.15 m for teleoperated trials, while for trials with the safety filter, all $h_\text{min}$ values are at least 0.05~m. This set of simulation evaluations demonstrates the efficacy and robustness of our proposed perceptive safety filter for guaranteeing safe teleoperation in different environments.

\renewcommand{\arraystretch}{1.2}
\begin{table}
    \centering
    \caption{Summary of Simulation Results}
\begin{tabular}{C{0.65cm}C{0.3cm}C{0.05cm}C{0.61cm}C{0.61cm}C{0.05cm}C{0.61cm}C{0.61cm}C{0.66cm}}
\hline\hline
\multirow{2}{*}{Env.} & \multirow{2}{*}{$d$} & & \multicolumn{2}{c}{Teleop Input} & & \multicolumn{3}{c}{Filtered Input} \\\cline{4-5}\cline{7-9}
&  & & $N_c$& $h_\text{min}$ & & $N_c$  & $h_\text{min}$ &$||\delta \bar{u}||$ \\\hline
 \multirow{3}{*}{1} 
& 28 && 83 & -0.22 && 0 & 0.05 & 0.03 \\
& 25 && 88 & -0.49 && 0 & 0.05 & 0.23 \\
& 22 && 63 & -0.50 && 0 & 0.05 & 1.00 \\\cline{1-9}
 \multirow{3}{*}{2} 
& 10 && 43 & -0.15 && 0 & 0.05 & 0.11 \\
& 13 && 34 & -0.15 && 0 & 0.05 & 0.07 \\
& 24 && 11 & -0.15 && 0 & 0.05 & 0.10 \\\cline{1-9}
 \multirow{3}{*}{3} 
& 27 && 52 & -0.15 && 0 & 0.05 & 0.04 \\
& 17 && 18 & -0.15 && 0 & 0.09 & 0.04 \\
& 18 && 19 & -0.19 && 0 & 0.05 & 0.04 
 \\\hline\hline
\end{tabular}
\label{tab:sim-result-summary}
\vspace{0.3em}\\[0.5em]
\justifying
\noindent Note: $d$ is the distance traversed by the robot (in meters), $N_c$ is the number of time steps with unsafe actions, $h_\text{min}$ is the minimum value of CBF attained (in meters), and $||\delta \bar{u}||$ is the average input adjustment applied by the safety filter (in meters). Visualizations of the environments are shown in Fig.~\ref{fig:sim-environments}.
\end{table}

\subsection{Experimental Results}

We performed experiments on a real MAV in order to showcase the applicability of our method in the real world.
We use an MAV based on the Holybro S500 quadcopter frame equipped with an Intel RealSense D455 RGB-D camera and an NVIDIA Jetson Orin 16 GB computer, which measures $0.8 \times 0.8 \times 0.4$~m.
All processing is performed on the MAV's onboard computer.
It is worth noting that no motion capture system or prior knowledge of the environment is used for the experiments. A video of the experimental results can be found at \href{http://tiny.cc/vi-slam-safe-filter}{http://tiny.cc/vi-slam-safe-filter}.

To evaluate the versatility of our proposed safety filter approach, the MAV is teleoperated to navigate in four distinct environment configurations. Visualizations of the test environments, overlaid with segments of the teleoperated and filtered inputs, are presented in Fig.~\ref{fig:experiment}. Subfigures (a)-(b) illustrate the case where the safety filter adjusts the teleoperated reference (depicted in red) to enable the MAV to safely traverse over obstacles while maintaining adequate clearance (shown in white). In subfigures (c)-(d), the teleoperated reference is positioned in close proximity to obstacles, and the filtered reference guides the MAV through narrow passages without collisions. Subfigures (e)-(g) and (h) respectively correspond to the scenarios where the MAV encounters the ground plane and is required to maneuver under tight constraints. Notably, in subfigure (h), the safety filter enables the MAV to successfully pass underneath a bridge without any collisions (Fig.~\ref{fig:cover}). These experimental results further validated the effectiveness of our safety filter in ensuring safe teleoperated MAV navigation across different scenarios and the applicability of our approach in a real-world setup where only onboard sensing and computation are accessible.

\section{Conclusion}
In this work, we proposed a perceptive safety filter framework that seamlessly integrates CBF certification with VI-SLAM and dense 3D occupancy mapping for safe teleoperated navigation of MAVs. Our system updates the environment map and the CBF in real time, solely relying on the sensors and computation resources available to the MAV onboard. Through a series of simulations and experiments, we demonstrated that the safety filter can efficaciously detect and correct teleoperated reference inputs that would otherwise lead to safety constraint violations, highlighting its effectiveness in ensuring the safe operation of MAVs in unstructured and cluttered environments. As future work, we plan to further leverage the proposed framework for semantic exploration and evaluate the system in outdoor environments.

\IEEEtriggeratref{15}
\bibliographystyle{IEEEtran}
\bibliography{references}
\end{document}